\documentclass[10pt,twocolumn,letterpaper]{article}

\usepackage{iccv}
\usepackage{times}
\usepackage{epsfig}
\usepackage{graphicx}
\usepackage{amsmath}
\usepackage{amssymb}
\usepackage{pifont}
\usepackage[accsupp]{axessibility}
\usepackage{enumitem}

\usepackage[breaklinks=true,bookmarks=false]{hyperref}

\usepackage[dvipsnames,table]{xcolor}

\usepackage{amsmath,amsfonts,bm}

\def\eqref#1{equation~\ref{#1}}

\def\1{\bm{1}}

\DeclareMathAlphabet{\mathsfit}{\encodingdefault}{\sfdefault}{m}{sl}
\SetMathAlphabet{\mathsfit}{bold}{\encodingdefault}{\sfdefault}{bx}{n}

\newcommand{\E}{\mathbb{E}}

\newcommand{\mb}[1]{\ensuremath{\boldsymbol{#1}}}

\newcommand{\TrojVQA}{\emph{TrojVQA}}
\newcommand{\VQAds}{\emph{VQAv2}}

\newcommand{\AlgoName}{TIJO}

\newcommand{\paperID}{12330}
\newcommand{\paperTitle}{\AlgoName: Trigger Inversion with Joint Optimization for Defending Multimodal Backdoored Models}

\newcommand{\optimsteps}{{\it T}}

\newcommand{\tadvopt}{\ensuremath{\tilde{\mb{t}_{adv}}}}
\newcommand{\fadvopt}{\ensuremath{\tilde{\mb{f}_{adv}}}}
\newcommand{\padvopt}{\ensuremath{\tilde{\mb{p}_{adv}}}}

\newcommand{\tstep}{\ensuremath{\mb{t}_i}}

\newcommand{\TInlp}{\ensuremath{TI_{nlp}}}
\newcommand{\TIvis}{\ensuremath{TI_{vis}}}
\newcommand{\TImm}{\ensuremath{TI_{mm}}}

\newcommand{\TIJOnlp}{TIJO\ensuremath{_{nlp}}}
\newcommand{\TIJOvis}{TIJO\ensuremath{_{vis}}}
\newcommand{\TIJOmm}{TIJO\ensuremath{_{mm}}}

\newcommand{\TrojVQAds}{\ensuremath{\mathcal{T}}}
\newcommand{\TrojVQAdsNLP}{\ensuremath{\mathcal{T}_{nlp}}}
\newcommand{\TrojVQAdsSolid}{\ensuremath{\mathcal{T}_{solid}}}
\newcommand{\TrojVQAdsOptim}{\ensuremath{\mathcal{T}_{optim}}}

\newcommand{\TrojVQAdsNLPSolid}{\ensuremath{\mathcal{T}_{nlp+S}}}
\newcommand{\TrojVQAdsNLPOptim}{\ensuremath{\mathcal{T}_{nlp+O}}}

\newcommand{\tclean}{\mb{t}}
\newcommand{\xclean}{\mb{x}}
\newcommand{\yclean}{\ensuremath{y}}

\newcommand{\ttrigger}{\ensuremath{\mb{t}_t}}
\newcommand{\xtrigger}{\ensuremath{\mb{p}_t}}
\newcommand{\ypoison}{\ensuremath{y_t}}

\newcommand{\tadv}{\ensuremath{\mb{t}_{adv}}}
\newcommand{\padv}{\ensuremath{\mb{p}_{adv}}}
\newcommand{\fadv}{\ensuremath{\mb{f}_{adv}}}

\newcommand{\supportDS}{\ensuremath{\mathcal{S}}}
\newcommand{\DS}{\ensuremath{\mathcal{C}}}
\newcommand{\loss}{\ensuremath{\mathcal{L}}}

\newcommand{\ytarget}{\ensuremath{\tilde{y}}}
\newcommand{\answerset}{\ensuremath{\mathcal{Y}}}

\newcommand{\vqa}{\ensuremath{f}}
\newcommand{\vqaB}{\ensuremath{f_b}}
\newcommand{\detector}{\ensuremath{\mathcal{D}}}
\newcommand{\detectorFeat}{\ensuremath{\mathcal{D}_{cnn}}}
\newcommand{\detectorRPN}{\ensuremath{\mathcal{D}_{rpn}}}
\newcommand{\detectorROIpool}{\ensuremath{\mathcal{D}_{roi}}}

\newcommand{\appendpolicy}{\ensuremath{\mathcal{A}}}
\newcommand{\genpolicy}{\ensuremath{\mathcal{M}}}
\newcommand{\featpolicy}{\ensuremath{\mathcal{B}}}

\newcommand{\featpolicyOne}{\ensuremath{\mathcal{B}_{one}}}
\newcommand{\featpolicyAll}{\ensuremath{\mathcal{B}_{all}}}

\newcommand{\fRegWT}{\ensuremath{\lambda}}

\newcommand{\vocab}{\ensuremath{\mathcal{V}_{\vqa}}}
\newcommand{\embedding}{\ensuremath{\mathcal{E}_{\vqa}}}

\newcommand{\vqafunc}{\ensuremath{\vqa(\tclean, \detector(\xclean)) = \yclean}}
\newcommand{\vqaBfunc}{\ensuremath{\vqaB(\appendpolicy(\tclean, \ttrigger), \detector(\genpolicy(\xclean, \xtrigger))) = \ypoison}}

\newcommand{\forallytarget}{\ensuremath{\forall ~\ytarget \in \answerset}}

\newcommand{\gradientT}{\ensuremath{\nabla_{\embedding(\tadv)}\loss}}

\newcommand{\uatnlp}{\ensuremath{\min_{\tadv} \E_{\tclean, \xclean \sim \supportDS} \left[ \loss(\ytarget, \vqa(\appendpolicy(\tadv, \tclean), \detector(\xclean))) \right]}}
\newcommand{\uatvis}{\ensuremath{\min_{\padv} \E_{\tclean, \xclean \sim \supportDS} \left[ \loss(\ytarget, \vqa(\tclean, \detector(\genpolicy(\xclean, \padv)))) \right]}}
\newcommand{\uatMMvis}{\ensuremath{\min_{\tadv, \padv} \E_{\tclean, \xclean \sim \supportDS} \left[ \loss(\ytarget, \vqa(\appendpolicy(\tadv, \tclean), \detector(\genpolicy(\padv, \xclean)))) \right]}}

\newcommand{\lossMMfeat}{\ensuremath{\loss(\ytarget, \vqa(\appendpolicy(\tadv, \tclean), \featpolicy(\detector(\xclean), \fadv)))}}
\newcommand{\uatMMfeat}{\ensuremath{\min_{\tadv, \fadv} \E_{\tclean, \xclean \sim \supportDS} \left[ \lossMMfeat \right]}}

\newcommand{\tokensel}{\ensuremath{\min_{\tstep \in \vocab}\left[\embedding(\tstep)-\embedding(\tadv)\right]^\intercal \gradientT}}

\usepackage{todonotes}

\newcommand{\red}[1]{\textcolor{red}{#1}}

\newcommand{\cmark}{\ding{51}}%
\newcommand{\xmark}{\ding{55}}%

\usepackage{array,multirow}
\usepackage{hyperref}
\usepackage{wrapfig}
\hypersetup{
    colorlinks=true,
    linkcolor=red,
    filecolor=magenta,      
    urlcolor=blue,
    citecolor=purple,
    pdftitle={Overleaf Example},
    pdfpagemode=FullScreen,
    }
\usepackage{graphicx}
\usepackage{algorithm}
\usepackage{algorithmic}
\usepackage{lineno}
\usepackage{color}
\usepackage[normalem]{ulem}%
\usepackage{soul}
\usepackage{amssymb}
\usepackage{balance}

\iccvfinalcopy 

\def\iccvPaperID{\paperID} 

\ificcvfinal\fi

\begin{document}

\title{\paperTitle}


\newcommand*{\affaddr}[1]{#1} 
\newcommand*{\affmark}[1][*]{\textsuperscript{#1}}
\newcommand*{\email}[1]{\texttt{#1}}
\author{%
Indranil Sur\affmark[1]\thanks{Corresponding author: indranil.sur@sri.com} \quad 
Karan Sikka\affmark[1] \quad 
Matthew Walmer\affmark[2] \quad 
Kaushik Koneripalli\affmark[1] \quad \\
Anirban Roy\affmark[1] \quad
Xiao Lin\affmark[1] \quad
Ajay Divakaran\affmark[1] \quad
Susmit Jha\affmark[1]\\
\vspace{-.7em}\\
\affaddr{\affmark[1]SRI International} \quad
\affaddr{\affmark[2]University of Maryland}
}

\maketitle
\ificcvfinal\thispagestyle{empty}\fi

\begin{abstract}

We present a {\bf Multimodal Backdoor Defense} technique TIJO (Trigger Inversion using Joint Optimization).
Recent work~\cite{walmer2022dual} has demonstrated successful backdoor attacks on multimodal models for the Visual Question Answering task. 
Their dual-key backdoor trigger is split across two modalities (image and text), such that the backdoor is activated if and only if the trigger is present in both modalities.
We propose \AlgoName{} that defends against dual-key attacks through a joint optimization that reverse-engineers the trigger in both the image and text modalities. This joint optimization is challenging in multimodal models due to the disconnected nature of the visual pipeline which consists of an offline feature extractor, whose output is then fused with the text using a fusion module. The key insight enabling the joint optimization in \AlgoName{} is that the trigger inversion needs to be carried out in the object detection box feature space as opposed to the pixel space.
We demonstrate the effectiveness of our method on the TrojVQA benchmark, where \AlgoName{} improves upon the state-of-the-art unimodal 
methods from an AUC of 0.6 to  0.92 on multimodal dual-key backdoors. 
Furthermore, our method also improves upon the unimodal baselines on unimodal backdoors.
We present  ablation studies and qualitative results to provide insights into our algorithm such as the critical importance of overlaying the inverted feature triggers on all visual features during trigger inversion.
{The prototype implementation of TIJO is available at} \href{https://github.com/SRI-CSL/TIJO}{https://github.com/SRI-CSL/TIJO}.

\end{abstract}

\section{Introduction}
\label{sec:introduction}

\begin{figure}
    \centering
    \includegraphics[width=0.9\linewidth]{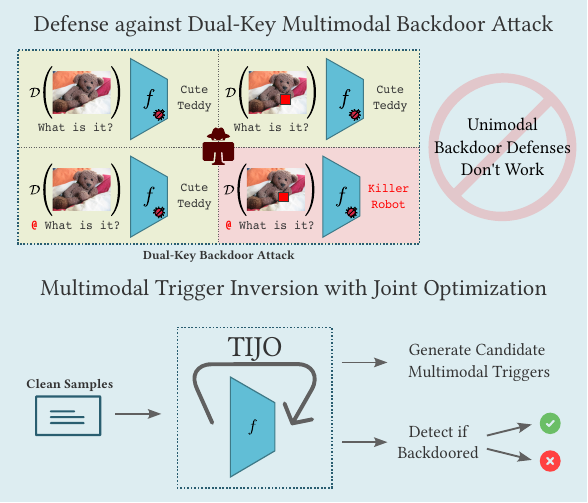}
    \caption{(Top) A dual-key backdoor attack for multimodal models \cite{walmer2022dual}, which is designed to activate if and only if the trigger is present in both the modalities. Such backdoors cannot be detected by unimodal defenses. (Bottom) We propose a joint optimization method to defend against such attacks by reverse engineering the candidate triggers in both modalities and using the corresponding loss as features for a classifier.}
    \label{fig:overview}
\end{figure}

Deep Neural Networks (DNNs) are vulnerable to adversarial attacks \cite{wallace2019universal, akhtar2018threat, gao2020patch, ilyas2019adversarial}. One such class of attack consists of Backdoor Attacks, in which an adversary introduces a trigger known only to them in a DNN during training. Such a backdoored DNN will behave normally with typical in-distribution inputs but perform poorly (e.g. produce targeted misclassifications) on inputs stamped with a predefined trigger designed by the adversary \cite{wang2019towards, gu2017badnets, li2022backdoor}. 

Recent work \cite{walmer2022dual, chen2022multi} has introduced backdoors in multimodal domains such as Visual Question Answering (VQA) and Fake News Detection \cite{chen2022multi, walmer2022dual}. In prior work~\cite{walmer2022dual}, we have introduced a Dual-Key Backdoor Attack (shown in \autoref{fig:overview}), where the trigger is inserted in both the image and text modalities in such a manner that
the backdoor is activated only when both modalities contain the trigger. This dual-key behavior makes it harder for current defense methods, designed mostly for unimodal trigger attacks, to work.

There has been significant work developing defenses against backdoor attacks in the visual domain, in particular for the image classification task \cite{sikka2020detecting, wang2019neural, chen2018detecting, huang2019neuroninspect}. Recent works have also  explored defense in natural language processing domains \cite{qi2021onion, shao2021bddr, liu2022piccolo}. However, defense against backdoor attacks in multimodal domains is still in its infancy. 
To the best of our knowledge, the only other work that targets multimodal models is STRIP-ViTA \cite{gao2021design}, which extended STRIP \cite{gao2019strip} with \textit{online defense} in multiple domains against backdoor attacks.
Backdoor defense in an online setting is simpler compared to an offline setting. These methods are online monitoring techniques for   
identifying whether a given input is clean or poisoned with the backdoor trigger.
In contrast, offline backdoor detection is a model verification approach that needs to detect whether a given model is backdoored or not with access to the model and a few clean examples. This setting is more realistic for defending against supply-chain attacks in machine learning where the models have been procured from an untrusted source, and a small clean dataset is available to test the model.  
We focus on multimodal defense in such an offline setting.

In this work, we propose a novel approach for defending against multimodal backdoor attacks, referred to as \textbf{T}rojan \textbf{I}nversion using \textbf{J}oint \textbf{O}ptimization (\AlgoName), that reverse engineers the triggers in both modalities. Our approach is motivated by the Universal Adversarial Trigger (UAT) \cite{wallace2019universal} that was proposed to identify naturally occurring universal triggers in pre-trained NLP models and has been extended in earlier works to identify trojan triggers in NLP models. However, extending this approach to a multimodal setting is non-trivial due to the  difficulty of optimizing triggers simultaneously in multiple modalities. Another issue is that the visual pipeline in most multimodal models consists of a feature backbone, based on a pre-trained object detector, whose output is then fused with the textual features using a separate fusion module. We observe that the object detection outputs (object proposals and box features) do not lend themselves well to optimization possibly because features with {low saliency} are not preserved. Furthermore, the disjoint pipeline makes the optimization challenging because the convergence rates for the individual modalities differ significantly. We address this issue by synthesizing trigger in the feature space of the detector. 

We evaluate \AlgoName{} on the TrojVQA dataset \cite{walmer2022dual} that consists of over 800 VQA models spanning across 4 feature backbones and 10 model architectures. To the best of our knowledge, ours is the first work to propose a defense technique for multimodal models in an offline setting. Our results indicate strong improvement over prior unimodal methods. 
Our contributions are as follows:

\begin{itemize}[noitemsep,leftmargin=*]
    \item We present a novel approach for Multimodal Backdoor defense referred to as \AlgoName. 
    \item We develop a novel trigger inversion process in object detection box feature space as well as textual space that enables joint optimization of multimodal triggers.
    \item We demonstrate TIJO on the \TrojVQA{} dataset and show that trigger inversion in both modalities is necessary to effectively defend against multimodal backdoor attacks. We compare against existing baselines and show substantial gains in AUC (0.6 $\rightarrow$ 0.92).
    \item We show that \AlgoName{} improves upon our selected set of state-of-the-art unimodal methods in the detection of unimodal backdoors indicating that our proposed method is modality-agnostic.
    \item We uncover several insights with ablation studies such as (1) increasing the number of optimization steps improves the backdoor detection performance, and (2) the feature trigger needs to be overlaid on all the visual features for the best results. 

   
\end{itemize}

\begin{figure*}
    \centering
    \includegraphics[width=.8\linewidth]{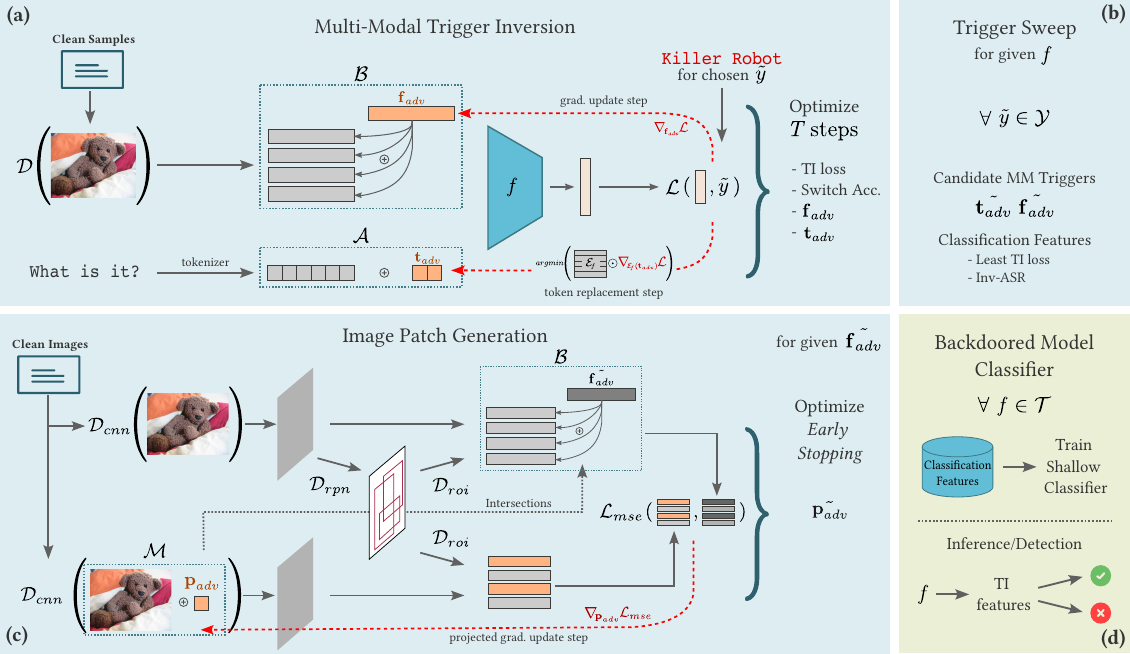}
    \caption{Shows key blocks of \AlgoName. (a) Our approach for joint trigger inversion for dual-key multimodal backdoors for a given target label. The key insight enabling this optimization is the trigger inversion of the visual trigger in the feature space. (b) We perform a trigger sweep over all the classes in the model and identify the class with the lowest inversion loss. (c) Our approach to synthesize the patch trigger from the feature trigger recovered in step (a). (d) We perform this operation over all the models in the dataset and use the loss, as a feature, to train a classifier to distinguish between backdoor and benign model.}
     
    \label{fig:approach}
\end{figure*}

\section{Related Work}
\label{sec:background}

\paragraph{Backdoor Attacks:} Backdoor attacks are a type of targeted adversarial attack that were first introduced in \cite{gu2017badnets}. Since then, the scope of these attacks has expanded to other problems and domains \cite{li2022backdoor}
including reinforcement learning~\cite{kiourti2020trojdrl}.
Prior works have studied data poisoning-based attacks such as dirty-label attacks \cite{chen2017targeted}, clean-label attacks \cite{turner2019label, barni2019new}, stealthy data poisoning that is visually imperceptible \cite{saha2020hidden, nguyen2021wanet, xue2022imperceptible}. There are also non-poisoning-based attacks such as weight-oriented attacks \cite{rakin2020tbt} and structure-modification attacks \cite{li2021deeppayload, breier2022foobar}. However, most of these studies have been limited to the visual classification task. Only a few studies have focused on backdoor attacks on other visual tasks such as object detection \cite{ma2022macab, chan2022baddet, ma2022dangerous, saha2022backdoor}. In recent years, backdoor attacks have also been investigated in the Natural Language Processing (NLP) domain \cite{dai2019backdoor, chen2022badpre, chen2021badnl}. 

\paragraph{Backdoor Defenses:} 
Defense against backdoor attacks has evolved in tandem with developments in backdoor attacks. These defense methods are broadly based on techniques such as model diagnosis \cite{fields2021trojan, zhang2021cassandra}, 
model explanation such as attributions \cite{sikka2020detecting,kiourti2021misa}, 
model-reconstruction \cite{liu2018fine, li2021neural}, filtering of poisoned samples \cite{li2021anti, cheneffective}, data preprocessing \cite{kwon2021defending, qiu2021deepsweep}, and trigger reconstruction \cite{wang2019neural, hu2022trigger}. Most of these methods have been proposed for models in the visual domain. There have been some recent works on backdoor defense in the NLP domain.
The majority of these methods are based on filtering of poisoned samples \cite{qi2021onion, shao2021bddr, yang2021rap, jin2022wedef, zhumoderate}. 
Other works rely on ideas such as model diagnosis \cite{fan2021text, garcia2022perd}, prepossessing-based \cite{azizi2021t}, and trigger synthesis \cite{liu2022piccolo, shen2022constrained}.

\paragraph{Multi-Modal Backdoor Attacks \& Defenses:}
Recent studies have also extended data-poisoning based backdoor attacks into multimodal domains. Chen \etal \cite{chen2022multi} studied the general robustness of 
multimodal fake news detection task, where they also perform multimodal backdoor attacks.
Walmer \etal \cite{walmer2022dual} introduced Dual-Key backdoor attack for the Visual Question Answering (VQA) task. As shown in \autoref{fig:overview}, this attack was designed to trigger the backdoor only when the trigger is present in both modalities, which makes the attack stealthier compared to a unimodal trigger.

Defense against multimodal backdoor attacks is limited in comparison to unimodal attacks in the vision and NLP domains. Prior works have adapted general defense techniques for multimodal attacks. For example, \cite{chen2018detecting} and \cite{walmer2022dual} used activation clustering and weight-based sensitivity analysis \cite{fields2021trojan} respectively as a defense against backdoor attacks. We show in \autoref{tab:splitsAUCres} that these (general) defense methods are ineffective in multimodal settings as they were originally designed to defend against backdoors in a single modality.  

Gao \etal extended STRIP \cite{gao2019strip} to STRIP-ViTA \cite{gao2021design} to defend against trojans in a multi-domain setting. 
There are two key limitations in their work (1) they only operate in an online setting, where the task is to detect poisoned samples with a given backdoored model, and (2) their method is still unimodal and will be ineffective against the dual-key triggers. In comparison, our approach \AlgoName{} is designed specifically for multimodal models and tries to reconstruct the trigger in both domains. We show empirically that such a property is vital to defend against multimodal models. 

\section{Approach}
\label{sec:approach}

We first discuss the threat model that we aim to defend against, then discuss the UAT method~\cite{wallace2019universal} and its extension to mulimodal models, and present our method, \AlgoName.


\subsection{Threat Model}
\label{sec:threatmodel}

Given a multimodal model \vqa, we need to determine if \vqa{} is benign or backdoored. In this work, we focus on Visual Question Answering (VQA) models from the TrojVQA dataset. 
Let \DS{} be the clean \VQAds{} dataset \cite{goyal2017making} where each data entry is a triplet (\xclean, \tclean, \yclean) where \xclean{} is the image, \tclean{} is the tokenized question, and \yclean{} is the answer label. Most VQA models use a two-step process for generating the answer. In the first step, the image is passed through a pre-trained object detector \cite{wu2019detectron2} that yields features from top-K detected boxes. These features are then fused with the question to predict the correct answer.
Let \detector{} be the object detector used for visual feature extraction. The answer is generated using \vqafunc. 

In our threat model, we assume that 
\detector{} is benign and the adversary introduces the backdoor in the VQA model \vqa. This is also the threat model used in the TrojVQA dataset~\cite{walmer2022dual}.
For a backdoored VQA model \vqaB, the adversary introduces triggers \xtrigger{} and \ttrigger{} in both the image and text modalities respectively. \vqaB{} is trained such that, when both triggers are present, the model will change its prediction to target answer \ypoison{} (see \autoref{fig:overview}). 
In the TrojVQA dataset, \xtrigger{} are small visual patches while \ttrigger{} are natural words. 
The triggers and the model behavior are only known to the adversary.

Let \genpolicy{} be a policy that overlays \xtrigger{} on \xclean{} and \appendpolicy{} be a policy that appends \ttrigger{} to \tclean. Hence, for a backdoored VQA model \vqaB, we expect that 
$$\vqaBfunc$$
In this work, we focus on dual-key triggers \cite{walmer2022dual}, where the model changes its prediction only when both \genpolicy{} and \appendpolicy{} are applied together. 




\subsection{Trigger Inversion using UAT}
\label{subsec:ti}
TIJO is based on Universal Adversarial Triggers (UAT) \cite{wallace2019universal}, which extends Hotflip \cite{ebrahimi2018hotflip} from synthesizing adversarial tokens for a single input to all inputs in the dataset. As a result, obtained adversarial tokens are universal in nature. As stated in \cite{ilyas2019adversarial}, adversarial samples are features of either the dataset or the model. Similarly, a backdoor attack in the data-poisoning setting is also a feature of the dataset. Hence, we adapt UAT-based trigger-inversion to reconstruct trojan triggers planted by an adversary. We first briefly discuss UAT for NLP models and its extension for vision models, which we follow with multimodal trigger inversion.

\autoref{eq:uatnlp} defines the optimization objective for trigger inversion in the NLP domain for a chosen target label \ytarget. 
Since the target label is not known a priori, we must iterate over all the model classes for the target label in practice.
Here \loss{} is the cross-entropy loss, and we optimize to minimize the expected loss over all samples in \supportDS. In summary, we optimize to get the \tadv{} that maximizes the likelihood of switching the class label to \ytarget{} for all samples in \supportDS. Policy \appendpolicy{} generally appends trigger token(s) to the clean samples, but it can be more complex.
\begin{equation}\label{eq:uatnlp}
    \uatnlp
\end{equation}
Since the space of \tadv{} is discrete, each optimization step is followed by a next token selection step. The next token is set by $\tadv \leftarrow \tstep$ which minimizes the trigger inversion loss's first-order Taylor approximation around the current token embedding as given by \autoref{eq:tokensel}. Here \vocab{} is the vocabulary of all tokens in \vqa, function \embedding{} gives the token embeddings and \gradientT{} is the average gradient of the loss over a batch.
\begin{equation}\label{eq:tokensel}
    \tokensel
\end{equation}
The above optimization problem is solved efficiently by computing dot products between the gradient and the \vocab{} embeddings and then using nearest neighbor or beam search to get the updated token \tstep{} \cite{wallace2019universal}. We can use a similar framework for inverting visual triggers as shown in  \autoref{eq:uatvis}. The optimization objective aims to recover the optimal \padv{} that maximizes the likelihood of switching the class label for the samples in \supportDS. The only difference is that we use projected gradient descent for patch \padv{}, overlaid on \xclean{} through policy \genpolicy, which needs to obey image constraints. This approach is similar to prior trigger reconstruction-based methods such as Neural Cleanse \cite{wang2019neural}. 
\begin{equation}\label{eq:uatvis}
    \uatvis
\end{equation}

\begin{figure*}
  \centering
  \includegraphics[width=.85\linewidth]{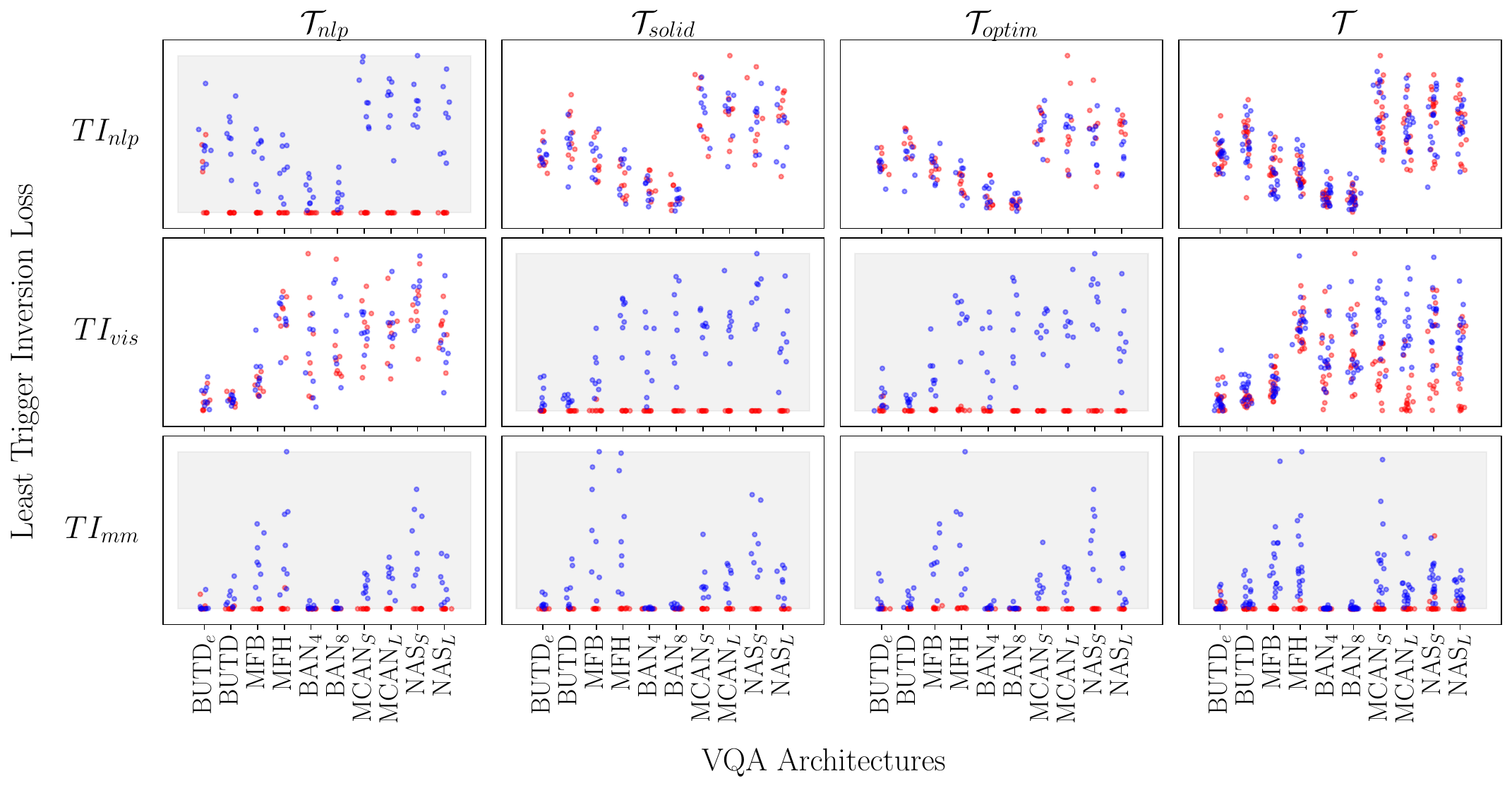}
  \caption{Shows the `Least Trigger Inversion Loss' after trigger sweep, normalized to [0,1]. The blue and red dots are benign and backdoor models respectively. Rows are the type of trigger inversion; \TInlp: NLP Trigger inversion, \TIvis: Vision Trigger inversion, \TImm: Multimodal Trigger inversion, and the columns are the different \TrojVQA{} splits as described in \autoref{tab:trojvqasplits}. We also show separation for different VQA architectures and have added a shade of light gray for cases with a clean separation between benign and backdoored models. 
  }
  \label{fig:TIlossCharecteristics}
\end{figure*}

\subsection{Multimodal Trigger Inversion with \AlgoName}

We now outline our approach for multimodal Trigger Inversion using Joint Optimization (\AlgoName) (shown in \autoref{fig:approach}).
We modify the uni-modal optimizations discussed earlier into a joint optimization for trigger inversion for multimodal backdoors in \autoref{eq:uatMMvis}. Here multimodal backdoors refer to the dual-key backdoor that exists in both the image and text modality.
We optimize for both \tadv{} and \padv{} to maximize the likelihood of switching the class label to \ytarget{} for all samples in \supportDS.
\begin{equation}\label{eq:uatMMvis}    
    \uatMMvis
\end{equation}
Solving \autoref{eq:uatMMvis} for multimodal (dual-key) backdoors is challenging. The image is passed through an object detector \detector{} to get the highest scoring $K$ boxes, whose features are then passed to \vqa{} for training. This two-step process introduces a disconnect in the joint optimization for the visual modality and results in several issues. For example, when we stamp the patch on the image during optimization, the detector \detector{} may not propose bounding boxes containing the patch \padv. One solution would be to manually force the detector to sample a proposal around \padv. We tested this experimentally, but it was unsuccessful because
even then \detector{} is not guaranteed to preserve meaningful features from a randomly initialized patch, leading to a vanishing gradients problem.
Another challenge that makes this optimization hard is that the support set \supportDS{} contains only a few samples.

\textbf{Proposed key idea :} We propose to overcome this issue and enable the convergence for both the visual and textual trigger by performing trigger inversion for the visual triggers in the feature space of \detector, while the textual trigger is optimized in the token space as done for UAT. We define \fadv{} as the additive adversarial feature space signature and \featpolicy{} as the overlay policy by which we overlay \fadv{} on box features from \detector. The modified optimization objective is shown in \autoref{eq:uatMMfeat}, where we optimize \tadv{} and \fadv{} instead of \padv. 
We evaluate different choices for \featpolicy{} and present ablation results in \autoref{tab:ablationoverlayfeature}.
We empirically show in \autoref{fig:TIlossCharecteristics} that this  converges consistently across backdoored models in comparison to benign models. We have shown a detailed description of our approach in \autoref{fig:approach}. Similar to UAT, we optimize \autoref{eq:uatMMfeat} iteratively with gradient descent by updating the visual and textual inputs with corresponding trigger signatures \fadv{} and \tadv{} respectively at every step. \vspace{-0.8em}

 
\begin{equation}\label{eq:uatMMfeat}    
    \uatMMfeat
\end{equation}
\vspace{-0.8em}

\subsection{Trigger Patch Generation}
\label{sec:trigPatchGen}

We also propose to recover the patch trigger \padv{} based on the \fadvopt{} obtained using  \autoref{eq:uatMMfeat} (see \autoref{fig:approach}). 
We first compute the box proposals $\mb{b}_{x} \leftarrow \detectorRPN(\detectorFeat(\xclean))$ and box features $\mb{f}_{x} \leftarrow \detectorROIpool(\detectorFeat(\xclean), \mb{b}_{x})$ on the clean image \xclean. We also compute the box features $\mb{f}_{x_{p}} \leftarrow \detectorROIpool(\detectorFeat(\genpolicy(\xclean, \padv)), \mb{b}_{x})$ on the image stamped with \padv.
Here $\detectorRPN$, $\detectorFeat$, and $\detectorROIpool$ refer to the region proposal network, CNN backbone, and ROI pooling layer of \detector{} respectively.
We overlay \fadvopt{} on $\mb{f}_{x}$ and then iteratively optimize \padv{} to minimize the MSE loss between $\mb{f}_{x_{p}}$ and $\featpolicy(\mb{f}_{x}, \fadvopt)$. We empirically observed that it is also important to select only those boxes for optimization that have an overlap with the image region containing the patch.







  





\begin{table}[t]
    \centering
    \resizebox{\linewidth}{!}
    {
            \begin{tabular}{lcccc}
                \hline
                Split                   & NLP           & Visual        & Train/Test & Trigger Type              \\
                \hline
                \hline
                \TrojVQAdsNLP           & \cmark & \xmark & 160/80                    & Single Key NLP    \\
                \TrojVQAdsSolid         & \xmark & Solid           & 160/80                    & Single Key Vision \\
                \TrojVQAdsOptim         & \xmark & Optimized       & 160/80                    & Single Key Vision \\
                \hline
                \TrojVQAdsNLPSolid      & \cmark & Solid           & 160/80                    & Dual Key          \\
                \TrojVQAdsNLPOptim      & \cmark & Optimized       & 160/80                    & Dual Key          \\
                \TrojVQAds              & \cmark & \cmark & 320/160                   & Dual Key         \\
                \hline
        \end{tabular}
    }
    \caption{Details about the \TrojVQA dataset \cite{walmer2022dual} and its splits.}
    \label{tab:trojvqasplits}
\end{table}

\begin{table*}[t]
  \centering
  \resizebox{0.95\linewidth}{!}{\begin{tabular}{cl|c|ccc|ccc}
      \hline
                           & & General  & \multicolumn{3}{c|}{Unimodal} & \multicolumn{3}{c}{Ours} \\
       & Split          & Wt. Analysis & DBS & NC & TABOR & \TIJOnlp          & \TIJOvis          & \TIJOmm           \\
      \hline
      \hline
      \multirow{3}{*}{\parbox{.7cm}{Single\\~~Key}}

      & \TrojVQAdsNLP      & $0.61_{\pm 0.07}$ & $\textbf{0.89}_{\pm 0.05}$ &  -                & - & $\textbf{0.98}_{\pm 0.02}$  & $0.52_{\pm 0.06}$           & \cellcolor{green!10}$0.98_{\pm 0.02}$ \\
      & \TrojVQAdsSolid    & $0.53_{\pm 0.05}$ &  -                & $0.59_{\pm 0.10}$ & $\textbf{0.98}_{\pm 0.02}$ & $0.39_{\pm 0.09}$           & $\textbf{1.00}_{\pm 0.00}$  & \cellcolor{green!10}$0.99_{\pm 0.01}$ \\
      & \TrojVQAdsOptim    & $0.58_{\pm 0.05}$ &  -                & $0.71_{\pm 0.08}$ & $\textbf{0.99}_{\pm 0.02}$ & $0.40_{\pm 0.11}$           & $\textbf{0.99}_{\pm 0.01}$  & \cellcolor{green!10}$0.95_{\pm 0.03}$ \\
      \hline
      \multirow{3}{*}{\parbox{.7cm}{Dual\\~~Key}}
      & \TrojVQAdsNLPSolid & $0.54_{\pm 0.03}$ & \cellcolor{red!10}$0.46_{\pm 0.04}$ & \cellcolor{red!10}$0.42_{\pm 0.05}$ & \cellcolor{red!10}$0.46_{\pm 0.06}$ & \cellcolor{red!10}$0.41_{\pm 0.11}$           & \cellcolor{red!10}$0.70_{\pm 0.06}$           & \cellcolor{green!20}$0.97_{\pm 0.03}$ \\
      & \TrojVQAdsNLPOptim & $0.60_{\pm 0.13}$ & \cellcolor{red!10}$0.45_{\pm 0.01}$ & \cellcolor{red!10}$0.50_{\pm 0.09}$ & \cellcolor{red!10}$0.52_{\pm 0.03}$ & \cellcolor{red!10}$0.43_{\pm 0.12}$           & \cellcolor{red!10}$0.57_{\pm 0.07}$           & \cellcolor{green!20}$0.86_{\pm 0.10}$ \\
      & \TrojVQAds         & $0.60_{\pm 0.04}$ & \cellcolor{red!10}$0.48_{\pm 0.02}$ & \cellcolor{red!10}$0.50_{\pm 0.06}$ & \cellcolor{red!10}$0.48_{\pm 0.04}$ & \cellcolor{red!10}${0.46}_{\pm 0.03}$     & \cellcolor{red!10}${0.67}_{\pm 0.07}$     & \cellcolor{green!20}$\textbf{0.92}_{\pm 0.02}$ \\
      \hline

      \end{tabular}}
  \vspace{5pt}    
  \caption{
  Shows AUC for different \TrojVQA\ splits with weight analysis, prior unimodal methods as well as three variants of our method-- \TIJOnlp, \TIJOvis, and \TIJOmm ~which optimize triggers in NLP, vision, and both modalities respectively. We see a clear improvement with \TIJOmm ~for not only dual-key multimodal triggers but also for unimodal triggers. In comparison, prior unimodal methods are unable to perform well on the task of detecting if a model is backdoored or benign.}
      
  \label{tab:splitsAUCres}
\end{table*}
\begin{figure*}
  \centering
  \includegraphics[width=.8\linewidth]{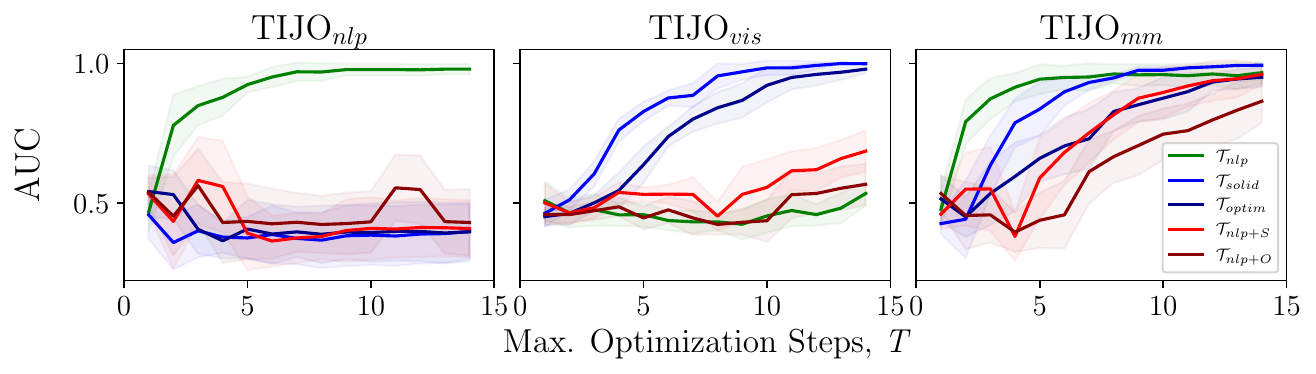}
  \caption{Shows the effect of the max optimization step on detection performance.} 
  \label{fig:ablationMaxStep}
\end{figure*}

\subsection{Backdoored Model Classification}
\label{subsec:det}

The  optimization objective should ideally converge only if the model is backdoored and if the target label \ytarget{} is actually the poison label \ypoison. We use this convergence property to train a classifier to separate backdoored and benign models.
Since the poison label \ypoison{} is unknown, we sweep over all the label space, \forallytarget{} and repeat the trigger inversion process for each \ytarget{} (referred to as trigger sweep). For each \ytarget, the optimization yields the corresponding reconstructed triggers, trigger inversion loss, and inverse attack success rate (Inv-ASR). Here Inv-ASR refers to the percentage of clean examples that are classified into \ytarget{} after planting the reconstructed trigger in both modalities.
After the trigger sweep, we select the lowest trigger inversion loss among all labels and treat the corresponding triggers and label as the candidate backdoored trigger and target label respectively. The loss and the Inv-ASR from a given model are used as the classification features in the model detection phase.

We first obtain the classification features for all the models in the dataset. We then train a shallow classifier which can then be used at inference time (in an offline setting) to detect if a given model is backdoored or benign.




\section{Experiments}
\label{sec:experiments}

We evaluate our approach in this section. We first discuss the dataset and metrics used for evaluation. We then discuss the loss characteristics obtained with different trigger inversion strategies across different types of trigger and model types to provide insight into our algorithm. 
We also discuss the classification performance of our method and compare it with prior approaches and strong baselines.
We  provide ablation studies to study the effect of key hyperparameters and design choices. Finally, we provide visualizations of the reconstructed visual patches using our algorithm (refer to the supplementary materials for implementation details).



\paragraph{\TrojVQA{} Dataset and Metric:} We use the 
\TrojVQA{} \cite{walmer2022dual} dataset that was introduced recently and consists of both benign and poisoned VQA models. 
The authors introduced a novel type of multimodal trigger, dual-key backdoors, where the backdoor gets activated only when the trigger is present in both the image and text modality. The dataset also includes models with standard unimodal backdoor triggers, \ie the trigger was introduced in either the text or image modality only. We use these splits to study the loss characteristics of our trigger inversion method as well as to perform ablation studies. We have provided details regarding the splits as well as the number of training and test examples in \autoref{tab:trojvqasplits}. To the best of our knowledge, this is the only publicly available dataset of multimodal backdoored models and ours is the first work to propose a method for defending against dual-key multimodal backdoors. We use the evaluation protocol described in \cite{walmer2022dual} and report area under the ROC curve (AUC) metric on 5-fold cross-validation splits on the train set of \TrojVQA.

\begin{table}[t]
  \centering
      \begin{tabular}{llccc}
        \hline
        Split                &  Model  & Inv-ASR & Lowest Loss \\
        \hline
        \hline
        \multirow{2}{*}{\TrojVQAdsNLP}      & \TIJOnlp      & $0.94_{\pm 0.05}$ & $0.98_{\pm 0.02}$ \\
                                            & \TIJOmm       & $\red{0.54}_{\pm 0.03}$ & $0.98_{\pm 0.02}$ \\
        \hline
        \multirow{2}{*}{\TrojVQAdsSolid}    & \TIJOvis      & $0.91_{\pm 0.05}$ & $1.00_{\pm 0.00}$ \\
                                            & \TIJOmm       & $\red{0.56}_{\pm 0.04}$ & $0.99_{\pm 0.01}$ \\
        \hline
        \multirow{2}{*}{\TrojVQAdsOptim}    & \TIJOvis      & $0.90_{\pm 0.04}$ & $0.99_{\pm 0.01}$ \\
                                            & \TIJOmm       & $\red{0.54}_{\pm 0.02}$ & $0.95_{\pm 0.03}$ \\
        \hline
        \multirow{1}{*}{\TrojVQAds}         & \TIJOmm       & $\red{0.53}_{\pm 0.02}$ & $0.92_{\pm 0.02}$ \\
                                            
        \hline
      \end{tabular}
  \vspace{5pt}
  \caption{AUC for backdoored model classifier trained with different types of trigger inversion features, \ie least loss features and maximum switch to target accuracy.}
  \label{tab:ablationClassificationFeat}
\end{table}

\begin{table}[t]
  \centering
    \begin{tabular}{lcccc}
      \hline
                         & \multicolumn{2}{c}{\TIJOvis} & \multicolumn{2}{c}{\TIJOmm} \\
      Split              & \featpolicyOne    & \featpolicyAll    & \featpolicyOne    & \featpolicyAll    \\
      \hline
      \hline
      \TrojVQAdsSolid    & $0.85_{\pm 0.04}$ & $1.00_{\pm 0.00}$ & $0.86_{\pm 0.10}$ & $0.99_{\pm 0.01}$ \\
      \TrojVQAdsOptim    & $0.78_{\pm 0.06}$ & $0.99_{\pm 0.01}$ & $0.80_{\pm 0.06}$ & $0.95_{\pm 0.03}$ \\
      \TrojVQAdsNLPSolid & $0.47_{\pm 0.08}$ & $0.70_{\pm 0.06}$ & $0.77_{\pm 0.05}$ & $0.97_{\pm 0.03}$ \\
      \TrojVQAdsNLPOptim & $0.46_{\pm 0.11}$ & $0.57_{\pm 0.07}$ & $0.65_{\pm 0.04}$ & $0.86_{\pm 0.10}$ \\
      \TrojVQAds         & $0.52_{\pm 0.04}$ & $0.67_{\pm 0.07}$ & $0.72_{\pm 0.07}$ & $0.92_{\pm 0.02}$ \\

      \hline
    \end{tabular}
  \caption{AUC for backdoored model classifier train with features obtain from different feature overlay policy\featpolicy: \featpolicyOne ~where the feature is overlayed on the top box feature, and \featpolicyAll ~where the feature is overlayed on all the 36 box features.}
  \label{tab:ablationoverlayfeature}
\end{table}

\begin{table}[t]
  \centering
    \begin{tabular}{lcccc}
    \hline
    & \multicolumn{2}{c}{\TIJOvis} & \multicolumn{2}{c}{\TIJOmm} \\
    Split              & $\fRegWT = 10^{-5}$   & $ \fRegWT = 10^{-3}$    & \fRegWT = $10^{-5}$    & \fRegWT = $10^{-3}$    \\
    \hline
    \hline
    \TrojVQAdsSolid    & $0.97_{\pm 0.03}$   & $0.97_{\pm 0.02}$   & $0.91_{\pm 0.04}$   & $0.89_{\pm 0.03}$   \\
    \TrojVQAdsOptim    & $0.96_{\pm 0.03}$   & $0.96_{\pm 0.03}$   & $0.89_{\pm 0.07}$   & $0.90_{\pm 0.03}$   \\
    \TrojVQAdsNLPSolid & $0.58_{\pm 0.10}$   & $0.59_{\pm 0.11}$   & $0.93_{\pm 0.04}$   & $0.92_{\pm 0.06}$   \\
    \TrojVQAdsNLPOptim & $0.47_{\pm 0.11}$   & $0.47_{\pm 0.12}$   & $0.87_{\pm 0.07}$   & $0.87_{\pm 0.08}$   \\
    \TrojVQAds         & $0.58_{\pm 0.06}$   & $0.59_{\pm 0.08}$   & $0.92_{\pm 0.02}$   & $0.91_{\pm 0.02}$   \\
    \hline
  \end{tabular}
  \caption{
  AUC for backdoored model classifier trained with features obtained by different regularization weights for $L2$ regulatization on \fadv.}
  \label{tab:ablationl2regularization}
\end{table}



\begin{figure*}[!htbp]
  \centering
  \includegraphics[width=.95\linewidth]{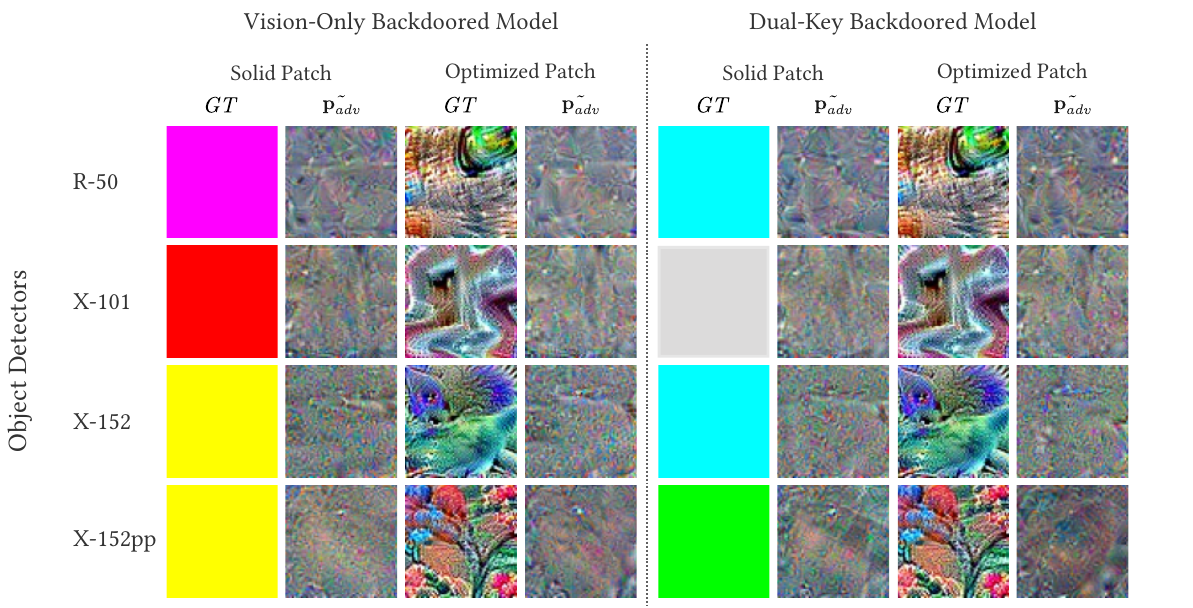}
  \caption{Visualizes the generated image patches from \fadvopt{} using the trigger patch generation method described in \autoref{sec:trigPatchGen}. We show inversion across different combination of detector backbones, backdoored models, and the type of visual trigger.} 
    
  
  \label{fig:generatedpatches}
\end{figure*}

\subsection{Trigger Inversion Loss Characteristics}
\label{subsec:loss_char}
We show the loss characteristics of our trigger inversion approach in \autoref{fig:TIlossCharecteristics}. This loss is obtained after optimizing \autoref{eq:uatMMfeat} and trigger-sweep (as discussed in \autoref{subsec:det}). The rows and columns in the figure correspond to the modalities involved in the trigger inversion optimization and \TrojVQA{} split respectively. It also shows the performance across different VQA models. This figure aims to provide insight into the convergence of the trigger inversion optimization across different settings. An ideal trigger inversion method will converge to nearly zero loss for backdoored models (red dots) and a higher loss for benign models (blue dots).

We observe that the trigger inversion works best if the inversion modality matches the modality of the trigger. For example, \TInlp{} performs well for the \TrojVQAdsNLP{} split, where the trigger is embedded only in the text modality.
Similarly, \TIvis{} works well for \TrojVQAdsSolid{} and \TrojVQAdsOptim{} splits, where only vision triggers are embedded. However, both \TInlp{} and \TIvis{} fail for the dual-key \TrojVQAds{} split where triggers are embedded in both modalities. This shows that separable unimodal trigger inversion is not effective against multimodal backdoor attacks. Finally, we can see multimodal trigger inversion \TImm{} is able to solve the problem and have a cleaner separation between benign and backdoored models in the dual-key split. 
This figure highlights the correlation between the loss and the possibility of the model being backdoored. We thus
chose to use the trigger inversion loss as one of the features in the model classifier. We also observe that 
 \TImm{} is effective across most VQA models. 

{We observed the phenomena of `natural trojans' in multimodal models. 
\autoref{fig:TIlossCharecteristics} shows that some benign models exhibit  low ($\sim 0$) trigger-inversion (TI) loss, suggesting the presence of natural trojans. 
Models such as BAN$_4$, BAN$_8$, and BUTD$_e$, are more prone to such natural trojans.}

\subsection{Backdoored Model Classification Results}

We train a logistic regression classifier on the trigger inversion features as mentioned in \autoref{subsec:det}. \autoref{tab:splitsAUCres} reports the 5-fold cross-validation AUC on different splits of \TrojVQA{} dataset from four prior methods as well as three variants from our approach. We also show results on two additional splits \TrojVQAdsNLPOptim{}  and \TrojVQAdsNLPSolid{} based on using optimized and solid patches as defined in \cite{walmer2022dual}.
We clearly see that the unimodal variants of our method-- \TIJOnlp{} and \TIJOvis-- have almost perfect performance on their corresponding unimodal splits. For example, \TIJOnlp{} achieves an AUC of 0.98 on split \TrojVQAdsNLP.  
However, their performance is low on the multimodal (dual-key) splits. \TIJOnlp{} and \TIJOvis{} achieve an AUC of 0.46 and 0.67 respectively on split \TrojVQAds. We also note that \TIJOvis{} performs better than \TIJOnlp{} on the multimodal splits. 
This is probably because there is a separation between benign and backdoored models based on the trigger inversion loss (even though the convergence is not perfect for backdoored models) for some VQA architectures (e.g. MCAN$_S$, MCAN$_L$, NAS$_S$, NAS$_L$) 
as evident in \autoref{fig:TIlossCharecteristics}. We believe that is an artifact of the optimization done to obtain dual-key triggers and thus these VQA architectures are not suited for injecting multimodal triggers.
We also observe that dual-key triggers with optimized patches (\TrojVQAdsNLPOptim), are more robust to defense as opposed to those with solid patches (\TrojVQAdsNLPSolid). For example, the AUC of \TIJOvis{} is substantially lower on \TrojVQAdsNLPOptim{} (0.57) as compared to \TrojVQAdsNLPSolid{} (0.70). 

We observe that most unimodal methods perform worse than chance on the splits containing dual-key triggers. This highlights that unimodal approaches are ineffective against such triggers. Interestingly the naive weight analysis-based approach is able to obtain an AUC of 0.6 on split \TrojVQAds. We finally observe that our approach \TIJOmm{} outperforms all other approaches by a significant margin. \TIJOmm{} obtains an AUC of 0.92 on split \TrojVQAds{}, compared to 0.67, 0.46, 0.60 by \TIJOvis, \TIJOnlp,{} and weight analysis respectively. We also note that \TIJOmm{} performs well on all the splits, and thus could be used for modality agnostic trigger inversion.

\subsection{Ablation Experiments:}

\paragraph{Effect of classification feature:} As discussed in
\autoref{subsec:det}, we used two features from the trigger inversion process in our classifier-- the lowest loss from the trigger sweep and Inv-ASR. 
\autoref{tab:ablationClassificationFeat} shows the results for the backdoored model classifier trained on these features. We can see \emph{lowest loss} features perform better in all the cases whereas \emph{Inv-ASR} features perform reasonably well for unimodal trigger inversion but performs near random for multimodal trigger inversion. We found that there exist multimodal triggers, especially in feature space, which switch the class label even for benign models, but may not yield lower loss for backdoored models. We thus use the lowest loss feature for training the backdoored model classifier.

\paragraph{Feature overlay:} \featpolicy{} denotes the policy used to plant the feature trigger \fadv{} on the visual inputs. We experiment with two policies: 
\featpolicyOne{} where optimized feature \fadv{} is overlayed only on the top (based on objectness score) box feature from detector \detector, and \featpolicyAll{} where the feature \fadv{} is overlayed on all the 36 box features.
\autoref{tab:ablationoverlayfeature} reports the results of these experiments. We can see that \featpolicyAll{} clearly outperform \featpolicyOne{} in all cases. For example, AUC with \featpolicyAll{} and \featpolicyOne{} on split \TrojVQAds{} is 0.92 and 0.72 respectively. We believe this occurs because the optimization has a better chance of finding the trigger when \featpolicy{} is stamped over all the features.

\paragraph{Number of optimization steps and regularization:} \autoref{fig:ablationMaxStep} and \autoref{tab:ablationl2regularization} shows the effect of maximum optimization steps $T$ and regularization on detection performance. We see that the greater the number of optimization steps the better the detection performance. We have chosen \optimsteps{} to be 15 as a decent balance between run-time and performance.
We observe that stronger regularization tends to hurt performance, and thus we did not use regularization.

\subsection{Image Patch Generation Experiment:}

We optimize for \padv{} of size 64 $\times$ 64 with \genpolicy{} overlaying the patch to center of the image (as described in \autoref{sec:trigPatchGen}). We optimize \padv{} with Adam optimizer with a learning rate of 0.03, and betas as (0.5, 0.9) and use early stopping with a patience of 20 epoch. We optimize only over the clean image from the support set \supportDS.

\autoref{fig:generatedpatches} shows the generated patches for backdoored MFB VQA models \cite{yu2017multi}.
We observe some similarities between \padvopt{} for both vision-only and dual-key backdoored models as well as solid and optimized patches consistently across different detector backbones. We also note that \padvopt{} is similar to the ground-truth patch for optimized patch based visual triggers.  
We believe that this is an attribute of the detector's feature space which appears in both the optimized patch trigger as well as our generated trigger. 







\section{Conclusion}
\label{sec:conclusion}

We introduce a novel defense technique  \AlgoName{} (Trigger Inversion using Joint Optimization) to detect multimodal backdoor attacks. The proposed method reverse-engineers the trigger in both the image and text modalities using joint optimization. Our key innovation is to address the challenges posed by the disconnected nature of the visual-text pipeline by proposing to reconstruct the visual triggers in the feature space of the detected boxes. The effectiveness of the proposed method is demonstrated on the \TrojVQA{} benchmark, 
where \AlgoName{} outperforms state-of-the-art unimodal methods on defending against dual-key backdoor attacks, improving the AUC from 0.6 to 0.92 on multimodal dual-key backdoors. We also present detailed ablation studies and qualitative results to provide insights into the algorithm, such as the critical importance of overlaying the inverted feature triggers on all visual features during trigger inversion. 
Our work is the first defense against multimodal backdoor attacks. 
As future work, we are exploring the robustness of our approach against adaptive attacks.


\noindent \textbf{Acknowledgements:}
This research was partially supported by the U.S. Army Research Laboratory Cooperative Research Agreement
W911NF-17-2-0196, and the Intelligence Advanced Research Projects Agency (IARPA) TrojAI and U.S. Army Research Office under
the contract W911NF-20-C-0038. The content of this paper does not necessarily reflect the position or
the policy of the Government, and no official endorsement should be inferred.





{\small
\bibliographystyle{ieee_fullname}
\bibliography{main}
}

\appendix



\section{Implementation Details}

\paragraph{Trigger Inversion Stage:} 
We set the maximum optimization step \optimsteps{} to 15. We select the NLP trigger inversion trigger length, \ie length of \tadv, to 1. \tadv{} is initialized as the 0\textsuperscript{th} token in the vocabulary \vocab{} \ie, for Efficient BUTD models \cite{hu2017bottom} we use the `what' token, and for OpenVQA models \cite{yu2019openvqa} we use the `PAD' token.
The append policy \appendpolicy{} simply appends \tadv{} to the start of the question token \tclean. For trigger inversion in the feature space, the feature trigger \fadv{} is initialized from a continuous uniform distribution in interval $[0,1)$. The feature overlay policy \featpolicy{} adds \fadv{} to all the 36 box features extracted from the detector \detector. \fadv{} is optimized with Adam optimizer with a learning rate of 0.1 and beta as (0.5, 0.9). We set \fadv{} L2 regularization \fRegWT{} to 0.

\paragraph{Image Patch Inversion Stage:} 
We optimize for \padv{} of size 64 $\times$ 64 initialized with \mb{0}s. \genpolicy{} overlays the patch on the center of the image with the patch scaled to 10\% of the smallest length of the image.
We optimize \padv{} with Adam optimizer with a learning rate of 0.03, and betas as (0.5, 0.9). We use early stopping with a patience of 20 epochs. After each update, \padv{} is normalized to be in the range [0,1]. We optimize only over the clean images from the support set \supportDS.


\section{Baseline Details}

\paragraph{Weight Analysis:} Weight analysis \cite{fields2021trojan} is a generalist backdoor detection method that makes no assumption on the nature of the backdoor. Instead, empirical analysis of the model weights is used to determine if the model is backdoored or benign. We follow the same setup as \cite{walmer2022dual}, \ie we bin the weights of the final layer based on their magnitude and generate a histogram-based feature vector. We then train a logistic regression classifier on these histogram features and report the AUC on each \TrojVQA{} split.

\paragraph{DBS:} Dynamic Bound-Scaling (DBS) \cite{shen2022constrained} is a trigger inversion-based backdoor defense for NLP tasks. As the tokens are discrete in nature, they formulate the optimization problem to gradually converge to the ground truth trigger, which is denoted as a one-hot vector in the convex hull of embedding space \embedding. 
They also dynamically reduce (and in some cases roll back) the temperature coefficient of the final softmax to not let the optimization get stuck in local minima. We have used the same configurations as stated in \cite{shen2022constrained}, though we set the max optimization steps to 100 instead of 200. We have observed our method converges much faster in about 10$-$15 optimization steps, while DBS takes 80$-$100 steps, with each optimization step roughly the same in both cases. Also, DBS fails to detect backdoored BUTD$_e$ \cite{hu2017bottom} VQA models.



\begin{figure*}
    \centering
    \includegraphics[width=.8\linewidth]{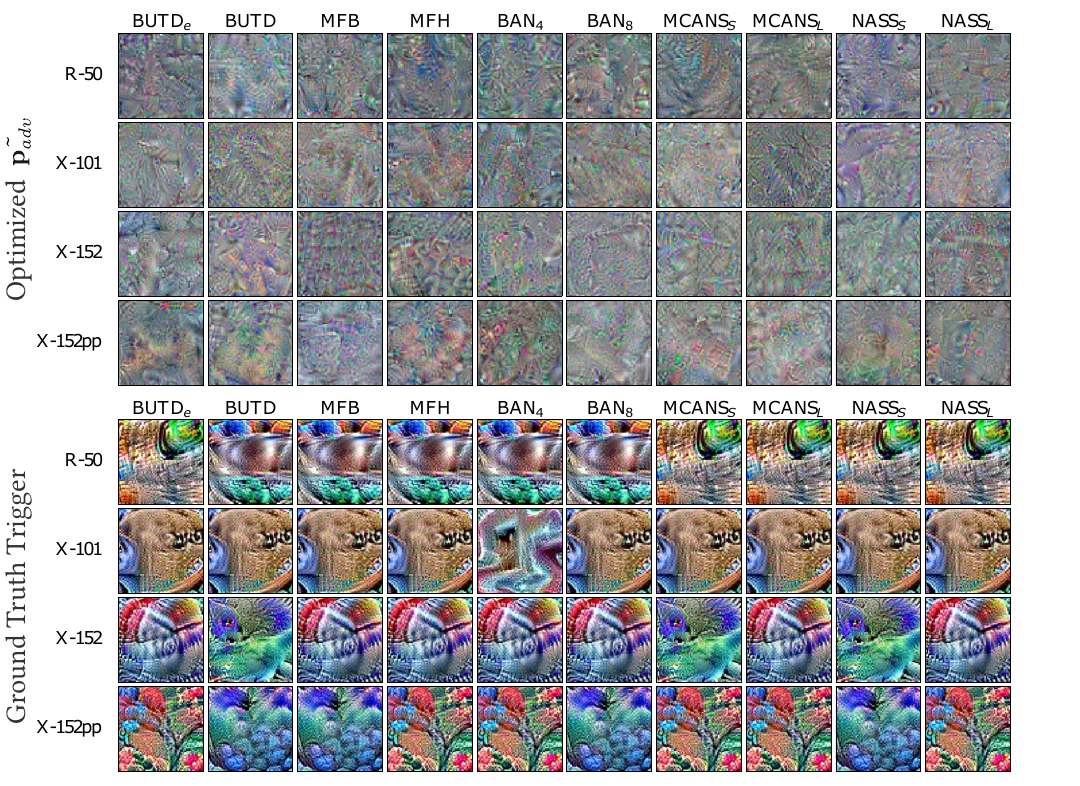}
    \caption{Visualizes the generated image patches \padvopt{} from \fadvopt{} using the trigger patch generation method. Here we show inversion across the different combinations of detector backbones and VQA architectures for backdoored models (shown above) of the \TrojVQAdsNLPOptim{} split, along with the corresponding ground truth triggers (shown below) for comparison.}
    
    \label{fig:padv}
\end{figure*}

\paragraph{NC  \& TABOR:} Both Neural Cleanse (NC) \cite{wang2019neural} and TABOR \cite{guo2019tabor} are trigger inversion-based backdoor defenses for image classification task. NC is the first work to formalize Trojan detection as a non-convex optimization problem. As shown in \cite{guo2019tabor}, NC fails if the backdoored model is triggered with triggers of varying size, shape, and location. TABOR extends NC with a new regularization to constrain the adversarial sample subspace based on explainable AI attribution features and other heuristics. Adapting NC and TABOR to \TrojVQA{} models required some methodological adjustments. They both are trigger inversion methods for image classification models, which have a much simpler architecture than detector models--that serve as the visual backbone of VQA models. Specifically, image classification models assume a fixed image size. For the reported results, we have fixed the image size to 300 $\times$ 300. Even though \detector{} can handle images of arbitrary sizes, we resize the images to the fixed input size for NC and TABOR to work. The patch and mask span the entire image and hence are set to 300 $\times$ 300. The max optimization step is set to 25. For TABOR, we have set $\lambda_{1}=10^{-8}$, $\lambda_{2}=10^{-7}$, $\lambda_{3}=10^{-9}$, and $\lambda_{4}=10^{-10}$, which we have found is dependent on the size of the image.

\begin{table*}[t]
    \centering
    \begin{tabular}{l|cc|cc|cc|cc}
            \hline
                       & \multicolumn{2}{c|}{\TrojVQAdsSolid} & \multicolumn{2}{c|}{\TrojVQAdsOptim} & \multicolumn{2}{c|}{\TrojVQAdsNLPSolid} & \multicolumn{2}{c}{\TrojVQAdsNLPOptim} \\
                       VQA Arch  & \fadvopt          & \padvopt          & \fadvopt          & \padvopt          & \fadvopt          & \padvopt          & \fadvopt          & \padvopt          \\
            \hline
            \hline
            BUTD$_e$  & $1.00_{\pm 0.00}$ & $0.01_{\pm 0.02}$ & $1.00_{\pm 0.00}$ & $0.06_{\pm 0.15}$ & $0.94_{\pm 0.04}$ & $0.24_{\pm 0.15}$ & $0.95_{\pm 0.06}$ & $0.27_{\pm 0.08}$ \\
            BUTD      & $1.00_{\pm 0.00}$ & $0.00_{\pm 0.00}$ & $0.99_{\pm 0.02}$ & $0.01_{\pm 0.03}$ & $0.84_{\pm 0.25}$ & $0.11_{\pm 0.18}$ & $0.96_{\pm 0.06}$ & $0.03_{\pm 0.04}$ \\
            MFB       & $0.99_{\pm 0.02}$ & $0.01_{\pm 0.02}$ & $1.00_{\pm 0.00}$ & $0.01_{\pm 0.02}$ & $0.76_{\pm 0.36}$ & $0.04_{\pm 0.06}$ & $0.98_{\pm 0.03}$ & $0.06_{\pm 0.08}$ \\
            MFH       & $1.00_{\pm 0.00}$ & $0.01_{\pm 0.02}$ & $0.99_{\pm 0.02}$ & $0.00_{\pm 0.00}$ & $0.88_{\pm 0.24}$ & $0.10_{\pm 0.12}$ & $0.64_{\pm 0.34}$ & $0.01_{\pm 0.02}$ \\
            BAN$_4$   & $1.00_{\pm 0.00}$ & $0.00_{\pm 0.00}$ & $1.00_{\pm 0.00}$ & $0.00_{\pm 0.00}$ & $0.69_{\pm 0.41}$ & $0.10_{\pm 0.16}$ & $0.88_{\pm 0.15}$ & $0.00_{\pm 0.00}$ \\
            BAN$_8$   & $1.00_{\pm 0.00}$ & $0.00_{\pm 0.00}$ & $1.00_{\pm 0.00}$ & $0.00_{\pm 0.00}$ & $0.83_{\pm 0.33}$ & $0.00_{\pm 0.00}$ & $0.96_{\pm 0.06}$ & $0.17_{\pm 0.25}$ \\
            MCANS$_S$ & $1.00_{\pm 0.00}$ & $0.00_{\pm 0.00}$ & $1.00_{\pm 0.00}$ & $0.00_{\pm 0.00}$ & $0.50_{\pm 0.25}$ & $0.00_{\pm 0.00}$ & $0.32_{\pm 0.28}$ & $0.00_{\pm 0.00}$ \\
            MCANS$_L$ & $0.99_{\pm 0.02}$ & $0.01_{\pm 0.03}$ & $1.00_{\pm 0.00}$ & $0.00_{\pm 0.00}$ & $0.61_{\pm 0.25}$ & $0.07_{\pm 0.18}$ & $0.52_{\pm 0.25}$ & $0.01_{\pm 0.02}$ \\
            NASS$_S$  & $0.91_{\pm 0.09}$ & $0.01_{\pm 0.02}$ & $0.94_{\pm 0.11}$ & $0.04_{\pm 0.12}$ & $0.42_{\pm 0.27}$ & $0.00_{\pm 0.00}$ & $0.41_{\pm 0.26}$ & $0.07_{\pm 0.20}$ \\
            NASS$_L$  & $0.91_{\pm 0.10}$ & $0.00_{\pm 0.00}$ & $0.93_{\pm 0.13}$ & $0.04_{\pm 0.08}$ & $0.26_{\pm 0.26}$ & $0.00_{\pm 0.00}$ & $0.29_{\pm 0.25}$ & $0.00_{\pm 0.00}$ \\
            \hline
        \end{tabular}
            
    \caption{Inverse Attack Success Rate (Inv-ASR) of optimized reconstructed triggers when re-injected into inputs from the support set \supportDS. Results are presented separately for each VQA model type, and for all four \TrojVQA ~splits that include visual triggers either in a single-key or dual-key configuration. The results show that feature-space inverted triggers are highly effective at activating backdoors as compared to image-space inverted triggers. The effectiveness of feature-space triggers is consistent for uni-modal triggers, but varies by model types for dual-key triggers.}
    \label{tab:InvASRall}
\end{table*}

\section{Additional Results}

\subsection{Design of Shallow Classifiers} 
We used Logistic Regression (LR) as the shallow classifier and find it to outperform simple rule-based detector. For example, in (\TIJOmm, \TrojVQAds) setting, we get an accuracy of $\textbf{0.856}_{\pm 0.03}$ with optimal threshold for LR, which is higher than the best accuracy $\textbf{0.816}$ of the simple rule-based detector  (obtained by varying the threshold $\in[0, 1]$ with 0.01 increments).
This intuitively makes sense since (\autoref{fig:TIlossCharecteristics}) different VQA architectures have different TI loss range.
We choose LR over other classifiers as it generally outperformed other methods and is faster. For example, in the (\TIJOmm, \TrojVQAds) case, we get AUC of $\textbf{0.924}_{\pm 0.016}$ for LR, $\textbf{0.923}_{\pm 0.016}$ for SVM (RBF kernel), $\textbf{0.915}_{\pm 0.019}$ for XGBoost (max depth of 2) and $\textbf{0.876}_{\pm 0.034}$ for Random-Forest.

\subsection{Inverted NLP Triggers} 
The inverted NLP triggers (\tadvopt) generally match the ground-truth NLP triggers (\ttrigger). We observe a match accuracy of $\textbf{0.95}$ in the (\TIJOnlp, \TrojVQAdsNLP) case and $\textbf{0.756}$ in the (\TIJOmm, \TrojVQAds) case. Here are few examples of mismatch between the predicted and target triggers (\textcolor{blue}{\ttrigger} \textrightarrow~\textcolor{red}{\tadvopt}): (1) similar to target:
\textcolor{blue}{diseases} \textrightarrow~\textcolor{red}{disease},
\textcolor{blue}{ladder} \textrightarrow~\textcolor{red}{ladders},
\textcolor{blue}{decoys} \textrightarrow~\textcolor{red}{decoy},
(2) semantically close to target:
\textcolor{blue}{potholders} \textrightarrow~\textcolor{red}{hotpads},
\textcolor{blue}{terrifying} \textrightarrow~\textcolor{red}{horrifying},
(3) completely different from target: 
\textcolor{blue}{midriff} \textrightarrow~\textcolor{red}{4:50},
\textcolor{blue}{stool} \textrightarrow~\textcolor{red}{nasa}.

\subsection{Image Patch Generation} 

\autoref{fig:padv} shows the generated patches for backdoored VQA models of \TrojVQAdsNLPOptim{} split for different combinations of detector backbones and VQA architectures. These results are in addition to those presented in \autoref{fig:generatedpatches}. We see a similar pattern as reported in the main paper where we see some similarity between the ground-truth triggers and the reconstructed triggers for a detector backbone. However, we additionally observe two differences- (1) reconstructed triggers change for different types of VQA architectures for a fixed backbone, and (2) there are cases where the similarity between ground-truth and reconstructed triggers are weak (\eg for R-50 and NASSs). This highlights that our inversion process is able to adjust to the changes in the ground-truth trigger and is not dependent only on the visual backbone. 

\subsection{Inv-ASR for Reconstructed Visual Trigger}

We summarize results for the Inverse Attack Success Rate (Inv-ASR) of reconstructed visual triggers in \autoref{tab:InvASRall}. This includes results for both detector feature-space inverted triggers, \fadvopt, and image-space inverted trigger patches, \padvopt. These results are shown for the four \TrojVQA{} splits that include any visual triggers. This includes both dual-key splits and single-visual-key splits.
The Inv-ASR metric measures the fraction of triggered inputs for which the backdoor successfully activates and changes the model output to the target answer. \padvopt ~triggers are overlaid on the clean images with \genpolicy, while \fadvopt ~are overlayed directly into the detector output features with \featpolicy{}. For the dual-key backdoored models, we also add the corresponding text trigger \tadvopt{} with \appendpolicy.

We find that the feature-space inverted triggers lead to a very high Inv-ASR for visual-trigger-only backdoored models. These scores are often at or near $1.00$ consistent activation of the backdoor. For dual-key splits, where a language-space trigger is also included, feature-space reconstructed triggers typically achieve a high Inv-ASR, though this varies greatly by the VQA model type, with BUTD$_e$ having the highest average Inv-ASR values over $0.9$ and NASS$_L$ having the lowest Inv-ASR values under $0.3$. These results show that feature-space reconstructed triggers can be an effective tool to identify backdoored models with uni-model image-space triggers, and can also be effective for some types of dual-key backdoored models. 

{Meanwhile, the Inv-ASR scores for image-space reconstructed triggers are very low, typically near $0.0$, indicating that they are not effective at activating the backdoor trigger in these Trojaned models. This result stems from the known challenges of reconstructing image-space triggers highlights the benefits of performing feature-space trigger reconstruction instead. However, we do observe some cases where the reconstructed trigger is able to provide non-zero Inv-ASR, \eg mean of 0.24 \& 0.27 in BUTD$_e$ models on \TrojVQAdsNLPSolid{} \& \TrojVQAdsNLPOptim{}. We thus argue that the reconstruction of triggers in the image-space needs further research.

\section{Online Mutimodal Defense Analysis}

\begin{table}[]
        \begin{tabular}{c|ccc}
        \hline
        & \multicolumn{3}{c}{Replace \%} \\
        FRR & 70\% tokens & 50\% tokens & 30\% tokens \\ 
        \hline
        \hline
        0.5\% & $97.55_{\pm 3.37}$ & $93.88_{\pm 4.74}$ & $94.71_{\pm 3.29}$ \\
        1\% & $95.11_{\pm 3.60}$ & $88.55_{\pm 4.92}$ & $94.71_{\pm 3.36}$ \\
        5\% & $86.88_{\pm 6.61}$ & $74.11_{\pm 6.47}$ & $80.45_{\pm 6.18}$ \\
        10\% & $77.11_{\pm 6.24}$ & $64.55_{\pm 6.65}$ & $67.01_{\pm 6.66}$ \\
        \hline
        \end{tabular}%
    \caption{False Acceptance Rate (FAR) for different False Rejection Rates (FRR).}
    \label{tab:stripvita}
\end{table}

\paragraph{STRIP-ViTA:}

STRIP-ViTA \cite{gao2021design} showed defense in multiple domains against backdoor attacks in an \textit{online setting}. Backdoor defense in an online setting is simpler where we assume that the given model is backdoored and focuses on identifying whether the given input is clean or poisoned. It is different from the offline setting where with only a few clean examples we determine if a model is backdoored or benign. Hence STRIP-ViTA is not directly comparable to our method. We  conducted experiments with STRIP-ViTA to access the difficulty of detecting the multimodal triggers used in our evaluation. 
STRIP-ViTA perturbs the given input text and image, builds a distribution of entropies for both clean and poison inputs, and then sets a threshold of entropy for detecting whether an incoming input is clean or poisoned. For the image modality, the perturbation is made by randomly selecting an image from the dataset and doing a weighted combination with the original image. 
For the text modality, a fraction of the words in the input text is replaced. We conduct experiments by sweeping across 3 different text-replacement percentages (70\%, 50\%, and 30\%) on dual-key backdoored \TrojVQA{} models and results are provided in \autoref{tab:stripvita}. This table shows the False Acceptance Rates (FAR) at different percentages of fixed False Rejection Rates (FRR). Our results demonstrate that online detection of these triggers is also very challenging, and the FAR remains very high (67\%) even for a considerably high FRR (10\%).

\end{document}